\documentclass{article} 
\usepackage{nips13submit_e,times}
\usepackage{hyperref}
\usepackage{url}
\usepackage{color}
\usepackage{xspace}
\usepackage{amsmath}
\usepackage{amssymb}
\usepackage{mdframed}
\usepackage[procnames]{listings}
\usepackage{graphicx}
\usepackage{tcolorbox}
\usepackage{color}
\usepackage{verbatim}
\definecolor{keywords}{RGB}{255,0,90}
\definecolor{comments}{RGB}{0,0,113}
\definecolor{red}{RGB}{160,0,0}
\definecolor{green}{RGB}{0,150,0}
\definecolor{codegray}{rgb}{0.5,0.5,0.5}
\usepackage[title]{appendix}

\usepackage{xcolor}

\title{Just ASK:  \\  Building an Architecture for Extensible Self-Service \\ Spoken Language Understanding }

\author{
Anjishnu Kumar \\
Amazon.com \\
\texttt{anjikum@amazon.com} \\
\And
Arpit Gupta \\
Amazon.com \\
\texttt{arpgup@amazon.com} \\
\And
Julian Chan \\
Amazon.com \\
\texttt{julichan@amazon.com} \\
\And
Sam Tucker \\
Amazon.com \\
\texttt{samtuck@amazon.com} \\
\And
Bjorn Hoffmeister \\
Amazon.com \\
\texttt{bjornh@amazon.com} \\
\And
Markus Dreyer \\
Amazon.com \\
\texttt{mddreyer@amazon.com} \\
\And
Stanislav Peshterliev \\
Amazon.com \\
\texttt{stanislp@amazon.com} \\
\And
Ankur Gandhe \\
Amazon.com \\
\texttt{aggandhe@amazon.com} \\
\And
Denis Filiminov \\
Amazon.com \\
\texttt{denf@amazon.com} \\
\And
Ariya Rastrow \\
Amazon.com \\
\texttt{arastrow@amazon.com} \\
\And
Christian Monson \\
Amazon.com \\
\texttt{cmonson@amazon.com}
\And
Agnika Kumar \\
Amazon.com \\
\texttt{agnika@amazon.com}
}

\nipsfinalcopy 

\begin{document}

\maketitle
\begin{abstract}

This paper presents the design of the machine learning architecture that underlies the Alexa Skills Kit (ASK) a large scale Spoken Language Understanding (SLU) Software Development Kit (SDK) that enables developers to extend the capabilities of Amazon's virtual assistant, Alexa.  At Amazon, the infrastructure powers over 25,000 skills deployed through the ASK, as well as AWS's Amazon Lex SLU Service. The ASK emphasizes flexibility, predictability and a rapid iteration cycle for third party developers. It imposes inductive biases that allow it to learn robust SLU models from extremely small and sparse datasets and, in doing so, removes significant barriers to entry for software developers and dialogue systems researchers.
\end{abstract}


\section{Introduction}
\label{sec:intro}

Amazon's \textit{Alexa} is a popular digital assistant that was not designed around a smart phone, but rather as a service for  \textit{ambient computing} \cite{sadri2011ambient} in the home. Due to the intuitiveness of the \textit{Voice User Interface (VUI)}, the demand from software engineering teams for voice controlled features far outstripped the pace at which the language experts designing Alexa's SLU system could accommodate them. Removing this dependency posed several challenges, the first was the need to build a self service SLU architecture that could transform natural language queries into API calls, while ensuring that the architecture played well with Alexa's existing user experience. This paper describes the design and creation of this service.

The Alexa Skills Kit (ASK) was initially designed to empower internal Amazon developers to prototype new features independently of Alexa's core Automatic Speech Recognition (ASR) and Natural Language Understanding (NLU) systems, and then extended to give third-party developers the same capabilities. In order to make Alexa a popular service for ambient computing, it was important to enable external developers to build sophisticated experiences for Alexa similar to the common operating systems of the smartphone era, Android and iOS.

\begin{figure}[h]
  \centering
  \includegraphics[width=.8\textwidth]{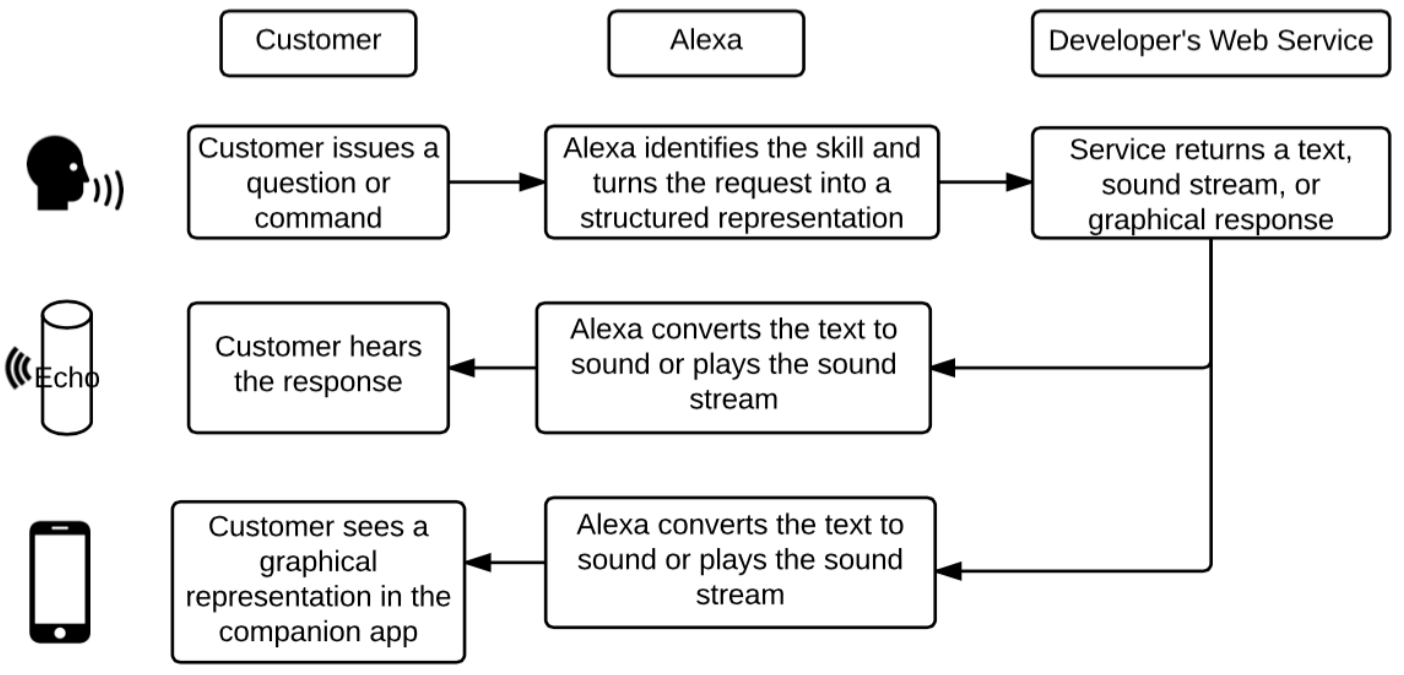}
  \caption{Interaction flow of an ASK skill}
  \label{alexa_interaction}
\end{figure}

An Alexa \textit{skill} is an SLU subsystem that has two customized components corresponding to the SLU stack - ASR and NLU. When the user speaks to a particular skill, the Alexa service handles the conversion of speech into text, performs intent classification, and slot-filling according to a schema defined by the skill developer. Besides writing the skill definition, the developer is also responsible for creating a web service, which interacts with JSON requests sent by Alexa.  Given the structured request from Alexa, the developer's web service can return text to be synthesized by Alexa's text-to-speech engine, an audio stream to be played back, with an optional graphical representation to be displayed in case the Alexa endpoint supports the visual modality. Figure \ref{alexa_interaction} illustrates the interaction flow of a skill. A common practice for Alexa skill developers is to use a \textit{serverless} endpoint such as AWS's Lambda product \cite{lambda2017}.

As of November 2017, there are over 25,000 Alexa Skills that have been built and deployed to customers via ASK. The SLU architecture described in this paper is also at the foundation of the Amazon Lex SLU service \cite{lex2017} and applications such as \textit{Amazon Connect}, a hybrid human/AI customer service product which uses Amazon Lex to power virtual agents. In the year following the launch of the ASK, competing SDKs following similar design principles have also been launched by Google and Microsoft, namely \textit{Actions on Google} and the \textit{Cortana Skills Kit}. 

ASK allows researchers to experiment on conversation and dialogue systems without the additional overhead of maintaining their own ASR and NLU systems \cite{lyons2016making, utama2017voice, serban2017deep}.

\section{Related Work}

Prior toolkits allowed users to build ASR and NLU models individually. For example, Kaldi \cite{povey2011kaldi} and HTK \cite{young1993htk} are popular toolkits for speech recognition. Stanford's Core NLP \cite{manning2014stanford} offers a suite of NLP libraries. These toolkits allow a lot of flexibility to the developer. For example, CoreNLP gives complete independence in selecting the libraries to use in the language understanding task. Similarly, Kaldi offers pre-built recipes which can be used as is or modified according to need of developer. However, this flexibility also poses challenges for the developer who is not well versed with Speech and NLP literature and ML methodology. These toolkits did not provide speech and language understanding capabilities in an open and self-service manner. At the other end of the spectrum, standards such as VoiceXML have existed since the early 2000s. They support clearly defined interfaces for software developers, but have supported only rigid command structures for end users. The need to develop a set of portable tools to address domain specificity with small datasets has long been considered a bottleneck for the large scale deployment of spoken language technology \cite{zue2000conversational}.

In parallel to our work, SpeakToIt (now DialogFlow) launched Api.ai and Microsoft launched LUIS.ai \cite{williams2015fast}, both of which provided self-service SDKs to third party developers to voice enable their applications in isolation. In our work, we attempt to close the gap by offering ASR and NLU models that work together out of the box with limited training samples and do not require expertise in either field, as well as the capability to rapidly deploy these systems directly to a large and growing audience by extending a widely available virtual assistant.

\section{Design Considerations}\label{sec:platform_design}
As an SDK for a voice assistant, speech recognition and language understanding technologies are the key value addition that Alexa can offer compared to existing frameworks. Since such a framework had never been built for a large user base before, we had to look beyond the software design tenets that were used to build largely monolithic high performance SLU systems in the past.  Successful modern software development frameworks such as Amazon Web Services, the Python programming language, and Linux, have large developer communities and could provide relevant design tenets. We describe some of these tenets below.

\textit{The system needs to offer modular and flexible building blocks.} To enable maximum flexibility, we chose to allow developers to specify commands they would want to support rather than limiting them to a set of commands or intents designed by Amazon. We chose to implement decoupled, modular subsystems that could be updated independently. We believe that decoupled systems are commonly studied in software engineering, but remain an underexplored area in existing machine learning research. In recent advances in deep learning research, there is a trend towards training complex end-to-end (E2E) models directly from data, \cite{Williams2017HybridCN, bahdanau2014neural, googlemt2016, serban2016building}. These models offer improved performance and are sometimes easier to maintain by reducing the number of components as compared to decomposed or cascaded architectures. However, joint modeling and end-to-end modeling can also introduce dependencies between constituent systems, and make it harder to deploy improvements independently \cite{creditcard}.

\textit{A third party must not be allowed to degrade the first party user experience.} The danger of customer experience degradation was mitigated by sandboxing skills and allowing them to be used only once  \textit{enabled}, either explicitly or implicitly. This design choice resulted in rapid adoption by developers since they were no longer in direct competition with each other or with the first party system, but made skill access more difficult for a customer by impairing the naturalness of the interaction.

\textit{The discoverable surface area for voice is limited, so the framework must prevent cybersquatting.} To prevent early developers from taking up valuable voice real-estate, a decision was made to allow overlapping skill names, and allowing developers to choose any name of their choosing as long as it does not reflect a brand protected by trademark or copyright which is not owned by them. Namespace disambiguation would be performed by the user at enablement time. Other frameworks chose to \textit{elect winners}, by having a more rigorous vetting process and by awarding valuable real estate to chosen vendors, which results in a more consistent user experience early on, but may limit innovation in the long run.

\textit{The framework should allow fast iteration to support a rapid VUI development cycle.} Since a skill developer did not have any usage data while developing a skill, enabling a rapid iteration cycle was important to enable them to quickly address issues. This requirement meant that SLU models needed to be trained in minutes, not hours, and created a tradeoff with the demand for highly accurate models.

\textit{The framework must remain relevant in the face of rapid advances in machine learning.} Since the system used machine learning extensively, the state of the art models implementing its underlying ASR and NLU technologies were changing rapidly.  The API contract thus had to be independent of the our initial model choice. We chose a lightweight shallow semantic parsing formalism comprised just of  \textit{intents} and \textit{slots}. This formalism is analogous to an API call, the \textit{intent} representing the function and \textit{slots} representing parameters to that function. The choice of this simple interface for communicating between the language understanding system and software developers meant that the underlying systems could be updated to use arbitrarily complex models and representations, which could then be compiled back into the API call representation. Since model architectures become outdated quickly, it was necessary to build up a repertoire of techniques that treat model choices as black boxes, in order to enable the rapid deployment of new model architectures without having cascading effects on other functionality. 

Since a developer could only provide a limited number of samples, any model trained on these utterances was unlikely to be of high quality. It thus became important to build components to leverage transfer learning and low resource learning techniques while remaining primarily model agnostic. 

This can be done in several ways. Firstly, \textit{data-based transfer}, that is, the curation of data resources that can be used generically. Secondly, \textit{representation-based} transfer, the creation of a label space that encapsulates prior knowledge about how different entities relate to each other. Thirdly, \textit{feature-based} transfer such as in \cite{daume2007frustratingly, Kim2015NewTL} or aided by unsupervised representation learning techniques \cite{pennington2014glove, mikolov2013distributed}. Investments in these transfer learning strategies are likely to remain relevant even when the state of the art model architecture changes. Once gains from generic techniques start to stagnate, it is imperative to invest in \textit{model-based} transfer learning strategies that exploit the specific characteristics of the machine-learning models being used \cite{46223, Kim2016FrustratinglyEN, trmal2010adaptation, rusu2016progressive}. In order to develop and deploy these strategies in a fast changing research landscape, it was critical to develop an infrastructure designed for the rapid deployment of research, we discuss these tradeoffs in subsequent sections.

\section{Customer Experience}\label{sec:developer_assets}

Customers primarily interact with skills primarily in 3 different ways.
\textit{Modal} interactions are explicit invocations where the customer first invokes a skill or service e.g. \textit{``Open Twitter"} then issues a command \textit{``Search for trending tweets"}. \textit{One-shot} invocation targets a skill or service and issues a command simulatenously, for example, \textit{"What is trending on twitter"}.

Both the modal and one-shot invocation modalities are supported by a combination of deterministic systems
and statistical shallow parsing models. The one-shot modality is only available for the set of skills a customer has \textit{enabled} previously,
in order to prevent unintended collisions. The explicit launch functionality, for example \textit{``Open the Twitter skill"}
attempts to disambiguate skills and do an implicit enablement if there is no ambiguity. 

A recently launched third modality is \textit{skill suggestion}. This allows for customers to be suggested a skill and routed to it when they issue a command that cannot be serviced by Alexa's first party systems but can likely be handled by a skill. This can be done by performing statistical matching between user utterances and
relevant skills by using techniques derived from information retrieval or using semantic matching 
performed by deep neural networks \cite{Kumar2017}.

Once a skill is invoked, a skill context is established, effectively sandboxing the interaction and preventing 
a customer from accessing any of Alexa's native capabilities until the skill exits,
either gracefully after task completion or because the customer ceases to interact with the device.

\section{BlueFlow: A Flexible Model Building Infrastructure}\label{sec:blue_flow}
Our modeling infrastructure is developed using \textit{BlueFlow}, a Python framework intended to accelerate the pace at which ML projects could be brought from research to production by using a shared codebase. We developed an internal design paradigm called \textit{CLI-to-Server}. The paradigm ensures that every API call defined by BlueFlow is also executable locally via a programmatically generated \textit{Command Line Interface (CLI)} in order to ensure that it is possible to reproduce the production stack during experimentation. Using autogenerated CLIs for the APIs adds an extra layer of abstraction to the codebase but helps mitigate a common machine learning anti-pattern in which production services and research tooling are implemented separately and go out of sync after a period of time.

BlueFlow is designed to enforce a clean separation between operational concerns and system logic using the constructs of \textit{artifacts, components, activities and recipes}. In a manner similar to some open source deep learning libraries \cite{chen2015mxnet},  BlueFlow uses python code constructs as a \textit{declarative language} to define a \textit{symbolic computational graph} for data flow management. This computational graph is a \textit{directed acyclic graph (DAG}) and can be serialized, then optimized and executed by a compatible \textit{executor} locally or in a distributed manner. Refer to Appendix \ref{sec:blueflow_arch} for details on the BlueFlow architecture and its syntax.

BlueFlow runs model building in production, but can also run it ad-hoc on a research cluster for conducting experiments. Having both research and production services use the same system allows us to quickly deploy new modeling techniques to production without spending significant time on productizing throwaway research code. The use of Python allows for rapid experimentation and a concise codebase, with an option to optimize bottlenecks in by using C++ bindings. However Python's type system increases reliance on unit test coverage. Along with the BlueFlow task execution framework, we extensively leverage technologies developed by AWS.  All static artifacts needed for a model build are stored in Amazon's Simple Storage Service (S3) \cite{s3} and all artifacts that are loaded at runtime are stored
in Amazon's highly performant key value store DynamoDB \cite{dynamodb}. 

\section{Skill Definition and Query Representation}\label{sec:alexa_ontology}
To create a new skill for Alexa, a developer begins by defining an \textit{interaction model}, which includes defining an intent schema, slot types, providing sample utterances corresponding to a simple grammar, and an invocation phrase. 


A wide variety of utilities exist to help developers define interaction model for an Alexa Skill, including but not limited to Skill Builder, a rich web interface with hints and completions and a testing console; AWS's Lex UI, which has an `export-as-skill' option; and programmatic skill management via the command-line using the Alexa Skill Management API - which also enables third party developers to start building skill development tooling.

ASK gives developers full freedom when defining a voice experience; however, a skill developer cannot realistically be expected to provide large and representative data samples that reflect real world usage. While this can be addressed in part by using transfer learning techniques, we also directly expose concepts from Alexa's internal domains in the form of \textit{builtin intents} and \textit{builtin slot types} for the developer to pick and choose from.

\begin{figure}[h]
  \centering
 \includegraphics[width=0.8\textwidth]{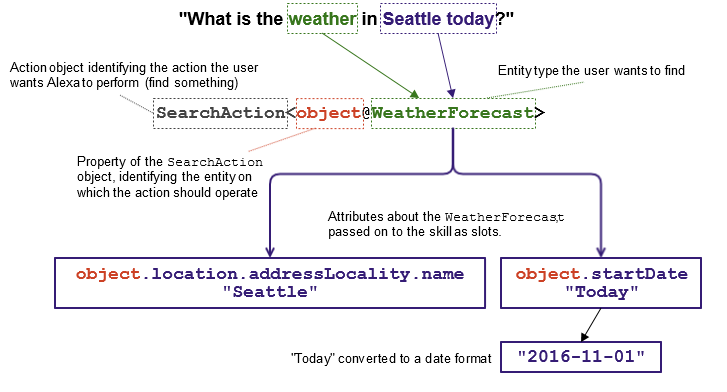}
 \caption{An utterance represented in AlexaMRL}
 \label{ontofig}
\end{figure}

\subsection{The Alexa Meaning Representation Language}

In this section we briefly describe the \textit{Alexa Meaning Representation Language (AlexaMRL)} \cite{Fan2017TransferLF}, which is necessary to understand builtin intents. The AlexaMRL is a decomposable semantic parsing formalism for defining spoken language commands, which allows for concepts to be efficiently reused across different SLU domains.

AlexaMRL is composed of \textit{Actions}, \textit{Entities}, and \textit{Properties}. Actions are a higher order abstraction than the \textit{intents} that are typically used in natural language tasks, and can be viewed as a templating system for intents. An Entity from the Alexa Ontology is analogous to a slot from an ASK developer's perspective. Properties are completely transparent to a skill developer, but under the hood, they tie Actions and Entities together by making certain Entity Types compatible with certain Actions. Entities possess Properties similar to object attributes in Object Oriented Programming. For example, \textit{LocalBusiness} is an Entity Type. It has Properties such as \textit{business hours, address, phone number} etc. Actions require Properties, i.e. they require for Entities to possess certain Properties in order to be compatible, a \textit{CallAction} cannot be completed unless it is operating on an Entity with a \textit{Callable} Property.

An intent represents an action the user wants to take. This could be searching or information, or playing a media object. For example, a \textit{FindPlaceIntent} intent would internally route to an API call for a location search engine. The reason an Action is more abstract than an NLU intent is because an intent needs to be aware of the surface forms (string representations) of the Entities it operates on, but an Action does not. Instead, an Action operates on abstract \textit{interfaces} that fulfill certain criteria. For example, An \textit{AddAction} requires an object Entity which can be added and a target Entity which can be added to. This is specified using the required Properties of the Action. AddAction has a required property \textit{targetCollection} which identifies the type of list to add to and an \textit{object} Property that marks the type of the object to be added.

\subsection{Exposing Query Representation to Skills }

Thus developers have an option to \textit{compile} builtin intents by filling an Action template with compatible Entity Types, which allows us to reuse data from Alexa's ontology via AlexaMRL. The skills that use compiled intents use shared deterministic models and efficiently reuse internal data resources for their stochastic models. We will discuss statistical modeling in greater detail in Section \ref{sec:model_building}.

Developers that choose to design their own intents with Alexa's builtin slot types, instead of Action templates, do not benefit from AlexaMRL's ability to automatically derive semantic roles, and cannot fully reuse shared deterministic models. In this case, only the shared data for an Entity Type can be reused, and the semantic roles of the Entities in an utterance must be derived from developer provided samples.

\section{Statistical Modelling}\label{sec:model_building}
In order for Alexa to turn user requests into structured representations, we need to build models based on the developer's definition for both the ASR and NLU systems. The first step to this process is to efficiently represent the grammars provided by the developer.

\subsection{Weighted Finite State Transducers}\label{sec:fst}

\begin{figure}[h]
  \centering
  \includegraphics[trim={0 0 0 0.1\textwidth}, width=1\textwidth]{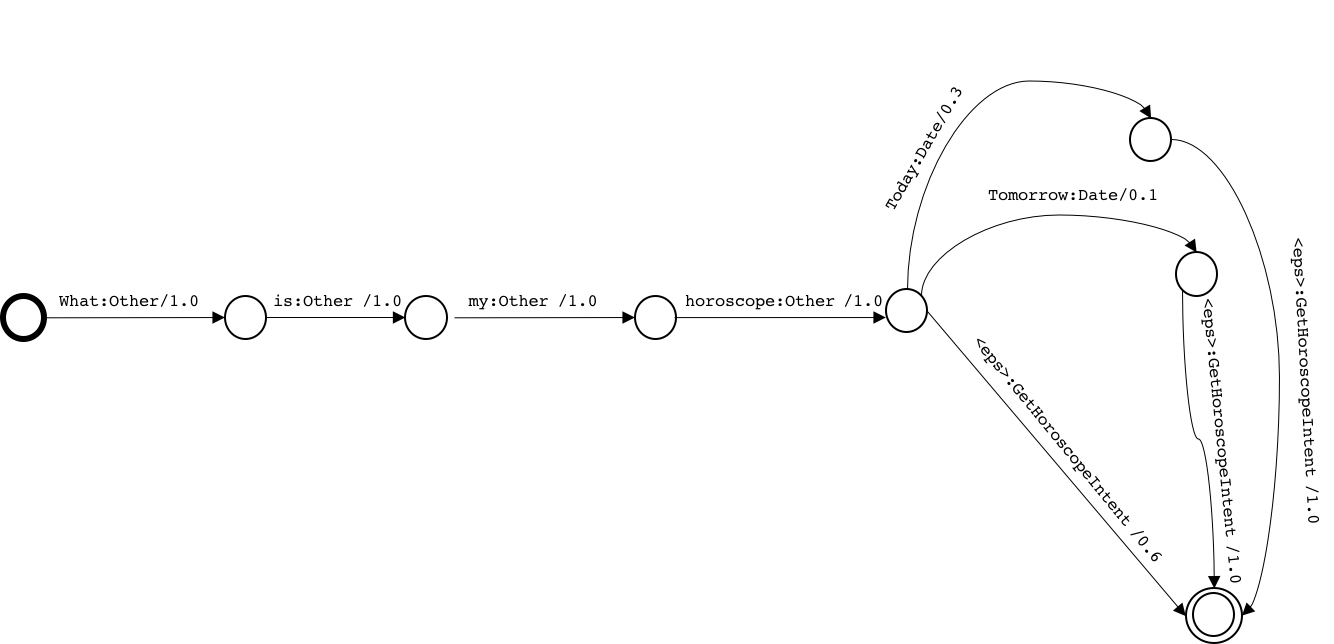}
  \caption{ wFST states are represented by circles and marked with their unique number. The input label \textit{i}, the output label \textit{o}, and weight \textit{w} of a transition are marked on the arcs. }
  \label{fstfig}
\end{figure}

Weighted Finite-State Transducers (wFST) provide an easy way to represent data under a weighted grammar. A path through the FST encodes an input string (or sequence) into output string. We generate FSTs for both ASR and NLU. ASR uses a skill specific FST to decode utterances into sentences defined by the developer, whereas NLU uses an FST to recognize intent and slot values. Both can be generated from the developer-provided sample utterances, intents, and slots. Most of the FST-generation code is shared between the ASR and NLU.  Fig. \ref{fstfig} shows an FST that recognizes the \textit{GetHoroscope} intent along with its \textit{Date} slot from an utterance.

This representation is powerful because we can impose arbitrary distributional priors on the grammar. We infer a distribution over intents and slots from the sample utterances. As the data in sample utterances is often imbalanced, we follow the \textit{principle of maximum entropy} to impose uniform priors over intents and then slots on the grammar. This is a configurable switch that can be turned off when the wFST is generated from the usage pattern of a skill.

The weighted FST representation can be used directly by common sequence models designed to work with text such as CRFs \cite{lafferty2001conditional} and LSTMs \cite{lample2016neural} by sampling utterances according to the distributional priors and feeding them into the respective models as training data via a data recombination technique similar to \cite{jia2016data}. One can also train directly on the lattice itself \cite{ladhak2016latticernn}.

\subsection{Automatic Speech Recognition}\label{sec:asr_recipe}

The ASR system uses weighted finite-state transducers (WFST) to represent the Hidden Markov Model (HMM) transducer (H), phone context-dependency transducer (C), pronunciation lexicon (L), and word-level grammar (G). These FSTs are composed to form an end-to-end recognition transducer. We refer the reader to Mohri et al.  \cite{mohri2002weighted} for details on these transducers and their role in ASR. The goal of the ASR recipe is to generate a word-level grammar (G) that guides the system to recognize utterances directed to the skill \cite{aleksic2015bringing}. The (G) decoder for ASK is a hybrid decoder that uses a skill-specific grammar as well as a main n-gram based Statistical Language Model (SLM) that shares data with other skills.
 
Continuous improvements are key to any machine learning product. In addition to regularly ingesting human-transcribed skill-directed utterances to the training corpus of the main SLM, we also ingest ASR recognition results directly as a form of semi-supervised training \cite{drugman2016active}. In order to reduce the risk of reinforcing recognition error, we employ weighted language model training \cite{Zhang2014KneserNeySO} where each semi-supervised sentence is weighted according to its confidence value given by the ASR system. These semi-supervised learning techniques ameliorate ASR errors due to distributional misallocations during the initial build phase.  

\subsection{Natural Language Understanding}\label{sec:nlu_recipe}

In this section we describe the Language Understanding component of the Alexa Skills Kit. Given an utterance, our NLU system aims to recognize its \textit{intent} and relevant parameters as \textit{slots}, a task known as shallow semantic parsing. The NLU system is divided into deterministic and stochastic subsystems.

The deterministic NLU subsystem uses FST data representation to compile sample utterances provided by developer into a recognizer model. This recognizer guarantees coverage on all utterances specified by a skill developer while designing her grammar, allowing for predictable experience for both the developer and the customer.

\begin{figure}[h]
  \centering
  \includegraphics[ width=.8\textwidth]{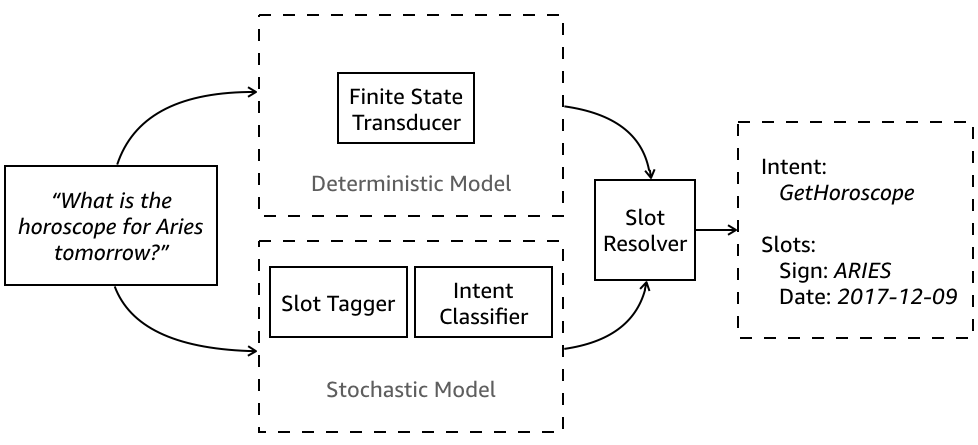}
  \caption{Overview of the hybrid NLU pipeline.}
  \label{fig:nlufig}
\end{figure}

The stochastic system uses BlueFlow to lend flexibility to the choice of model. We have built individual algorithmic components which implement a linear chain CRF \cite{lafferty_conditional_2001}, and a maximum entropy classifier \cite{berger_maximum_1996}, an updated pipeline using LSTMs  for each task \cite{yao2014spoken} and joint neural models for entity and intent prediction \cite{liu2016attention, hakkani2016multi}. This allowed us to experiment with different configurations as long as it conforms to the API of receiving a sentence and returning a semantic frame. In fig \ref{fig:nlufig} we show a traditional NLU system which performs entity recognition followed by intent classification and finally slot resolution.

\setlength\belowcaptionskip{-3ex}

\subsubsection{Knowledge Injection}

The performance of statistical models can be greatly improved by using features derived from Knowledge-Base (KB) lookups, but it is infeasible to perform expensive queries at runtime. By using efficient data structures to encode sketches of relevant portions of a knowledge base, we can make them available to statistical models as feature extractors. This is done by encoding ontologically derived word-clusters as bloom filters during training time \cite{liu2016personalized, smallffn}, including those from custom types defined by developers. These features can be extremely useful for training models in a low data regime since they encourage feature generalization to classes or categories, with a small probability of false positives. This can massively increase the effective vocabulary of a small model without blowing up its feature-space, and has added benefit of enabling class coverage improvements to deploy asynchronously without needing to retrain statistical models.

Training with KB features can lead to language feature under-training, the models learn to rely too heavily on KB features to distinguish entities. This can be addressed by introducing a \textit{knowledge dropout} regularization parameter to prevent the model from overfitting \cite{yang2016drop}.

\subsubsection{Model Optimization for Deployment}

Skill models are stored in DynamoDB \cite{dynamodb} and loaded at runtime, and network calls account for the bulk of added latency. We use standard techniques like feature hashing \cite{weinberger2009feature}, weight quantization, \cite{joulin2017bag} and sparsity constraints via modified elastic net regularization \cite{zou2005regularization} to ensure that our models are small enough to be stored cheaply and can be transferred over network calls quickly with no statistically significant loss in accuracy.

\section{Dialogue Subroutines}
ASK supports the specification of dialogue subroutines for common tasks, 
a unified dialogue model automates simple procedural dialogue capabilities,
 such as slot elicitation (e.g., \textit{User: Get me a
  cab -- Alexa: Where do you want to go?}) confirmation questions
(e.g., \textit{User: Portland -- Alexa: You want to go to Portland,
  Oregon, right?}) and other easily specified mechanisms. These procedural subroutines can either 
  be invoked by developers to issue a \textit{dialogue act directive} during the course of interaction, or defined as part of the interaction model. Although this system fulfills many usecases, we recognize the limitations of having a simple dialogue system.
 
\section{Conclusions and Future Work}\label{sec:conclusion}
We described the ASK SLU service which allows third party developers to expand the capability of the Alexa Voice Service. Our system abstracts away the intricacies of ASR and NLU, and provides developers with an interface based on structured request data.  The ASK service now hosts over 25,000 consumer facing SLU subsystems that improve over time, with hundreds of new Skills being added every week. This implies that we have made significant progress towards our goal of building a fully extensible SLU architecture. However challenges still remain in building a seamless user experience, and to enable the creation of useful and engaging agents. In order to seed research in this area, Amazon has introduced the Alexa Prize, a 2.5 million dollar university competition to advance conversational AI through voice. Alexa prize bots were launched to customers as a skill using ASK, and early results have shown promise \cite{serban2017deep}.

\bibliographystyle{IEEEbib}
\bibliography{ask}

\newpage
\begin{appendices}

\section{BlueFlow Architecture}
\label{sec:blueflow_arch}
\textbf{Architecture.} The fundamental units of BlueFlow are \textit{components}. Component are wrapped in \textit{activities}. Activities are chained together into \textit{recipes}. Recipes are executed by an \textit{executor}. 

\textbf{Components.} A component is a function that solves a defined task. This can be simple (convert from format A to format B) or complex (train a classifier). In either case, a component is the smallest unit that other scientists would expect to reuse. Components are normal Python code and can be used independently of BlueFlow. 

\textbf{Activities.} Activities are lightweight wrappers around components, responsible for fetching data from \textit{artifacts} and then running a component. Artifacts are lazy file like objects that support a uniform interface for reading and writing data from heterogeneous data sources such as local files, DynamoDB \cite{dynamodb} and S3 \cite{s3}. Activities are expressed as regular Python, but they specify input and output via an annotation syntax.

Below we show a simple activity to train a Maximum Entropy Classifier.

\lstset{language=Python, 
        basicstyle=\ttfamily\footnotesize, 
        keywordstyle=\color{keywords},
        commentstyle=\color{comments},
        stringstyle=\color{red},
        showstringspaces=false,
        procnamekeys={def,class},
        tabsize=2,
        xleftmargin=.0\textwidth,
        numberstyle=\tiny\color{codegray},
        label=activity}

\begin{lstlisting}
@Activity(inputs=('features_artifact'), 
                outputs=('model_artifact'))
def train_classifier(features_artifact, model_artifact):
  model_file = components.train_maxent(features_artifact.fetch())
  model_artifact.put(model_file)
\end{lstlisting}

\textbf{Recipes.} A recipe is a chained together set of activities. Recipe code is no longer normal Python, but rather uses Python language constructs to define a symbolic graph of data flowing through Activities as Artifacts, and can be serialized as a Directed Acyclic Graph (DAG). Then, we can execute the independent paths of the DAG in parallel either on the same machine or on a cluster.

Below we show a recipe for training a Maximum Entropy Classifier.

\lstset{language=Python, 
        basicstyle=\ttfamily\footnotesize, 
        keywordstyle=\color{keywords},
        commentstyle=\color{comments},
        stringstyle=\color{red},
        showstringspaces=false,
        procnamekeys={def,class},
        tabsize=2,
        label=recipe}
\begin{lstlisting}
@Recipe
def build_ic_model(data_file, executor):
    data_artifact = executor.new_artifact(data_file)
    features_artifact = executor.new_artifact()
    model_artifact = DynamoDBArtifact('models/classifier')
    extract_features(data_artifact, features_artifact)
    train_classifier(features_artifact, model_artifact)

\end{lstlisting}

\textbf{Executors.} An executor abstracts the details of a recipe execution.
It is responsible for vending appropriate artifacts and for taking the appropriate actions when executing specific activities.
Thus, a \textit{local} executor vends local files as intermediate artifacts and runs recipe's activities on the same machine, whereas a \textit{remote} executor vends S3 files as intermediate artifacts and runs recipe's activities on a cluster.

The executor is also responsible for performing recipe optimizations, via improved scheduling, IO or smart object caching. The local executor can be used for low latency model builds such as the ASK model builds. For offline experiments with higher throughput requirements we use the remote executor to run jobs on a cluster. Switching from a naive local executor to a multithreaded local executor results in model build speed increases of about 50 percent.

Other than recipe execution, executors add important features for production code such as logging, metrics, and retries, that would otherwise result in significant boilerplate.
Every Recipe is automatically converted into an equivalent command line tool for local execution and reproduction.

\section{Intent Schema and Custom Slot Types}\label{sec:schema}

The \textit{intent schema} specifies the intents supported by a skill and the expected slots for an intent.

An \textit{intent} is the intended action for an utterance. Developers may define their own intents or use the Amazon provided built-in intents. An intent may accept a \textit{slot} as an argument to represent an entity along with it's semantic role in the sentence. The data type of the slot is captured by the \textit{slot type}, which corresponds to entity types from the ontology.

Developers may use the Amazon provided built-in slot types or define their own \textit{custom slot types}. Each custom slot type requires a list of representative values. 
\lstdefinelanguage{json}{
        basicstyle=\ttfamily\footnotesize, 
        showstringspaces=false,
        tabsize=2,
        numbers=left,
        numberstyle=\tiny\color{codegray},
        caption={Sample Intent Schema for a Horoscope skill from the ASK Documentation},
        breaklines=true
}

\begin{lstlisting}[language=json,firstnumber=1]
{
  "intents": [
    {
      "intent": "GetHoroscope",
      "slots": [
        {
          "name": "Sign",
          "type": "ZODIAC_SIGNS"
        },
        {
          "name": "Date",
          "type": "AMAZON.DATE"
        }
      ]
    }
  ]
}
\end{lstlisting}

\textit{Sample utterances} form the training data used for building spoken language understanding models. Each sample is labelled with an intent followed by an utterance corresponding to the intent. 

\lstdefinelanguage{utterance_format}{
        basicstyle=\ttfamily\footnotesize, 
        showstringspaces=false,
        numbers=left,
        numberstyle=\tiny\color{codegray},
        caption={Sample Utterances for a Horoscope skill},
        breaklines=true
}
\begin{lstlisting}[language=utterance_format,firstnumber=1]
GetHoroscope what is the horoscope for {Sign}
GetHoroscope what will the horoscope for {Sign} be on {Date}
GetHoroscope get me my horoscope
GetHoroscope {Sign}
\end{lstlisting}

\end{appendices}


\end{document}